\ifdefined\XeTeXversion\else\pdfoutput=1\fi
\documentclass[11pt]{article}

\PassOptionsToPackage{table}{xcolor}
\usepackage[preprint]{acl}

\usepackage{times,latexsym}
\usepackage[T1]{fontenc}
\usepackage[utf8]{inputenc}
\usepackage{amsmath,amssymb}
\usepackage{microtype}
\usepackage{inconsolata}
\usepackage{graphicx,booktabs,colortbl,array,pifont}
\usepackage{enumitem}

\title{Evaluating Modeling Approaches for Experience-Level Classification in Job Description}

\author{Celia Liang, Eddie Wu, Shiqi Wang, Yonah You \\
  New York University\\
  Computer Science Department \\
  \texttt{cl7093@nyu.edu, mw5068@nyu.edu, sw6423@nyu.edu, xy2382@nyu.edu}}

\begin{document}
\maketitle
\begin{abstract}
This paper investigates the task of predicting job experience levels in recruitment texts, aiming to automatically identify the qualifications required for positions. Unlike traditional text classification, recruitment texts typically possess explicit internal structures, with different paragraphs playing disproportionate roles in conveying experience clues. To address this, we propose a structure-aware Section-Aware BERT approach that segments and encodes key paragraphs (titles, responsibilities, requirements) for integrated modeling, building upon rule-based systems and classical baselines TF-IDF. Simultaneously, we evaluate large language models under both few-shot and fine-tuning settings on the same dataset to compare the capability boundaries of different modeling paradigms. Experimental results demonstrate that explicitly leveraging text structure significantly improves experience level prediction performance, particularly in scenarios with ambiguous job titles. Further error analysis reveals systemic challenges in this task, including confusion between Entry and Senior levels and the blurred boundaries of Mid-level positions. This research provides an effective modeling approach and analytical framework for understanding structured recruitment texts.
\end{abstract}

\noindent\textbf{Keywords:} Recruitment Text Analysis, Job Descriptions, Experience Level Prediction, Section-Aware BERT

\section{Introduction}\label{introduction}

In recent years, job recruitment text analysis has emerged as a significant research direction within the field of natural language processing and has played a central role in applications such as job recommendation systems and labor market forecasting. Among these, experience level classification stands as a critical sub-task, holding great importance for automatically understanding job qualification requirements and enhancing the quality of job-candidate matching. However, unlike traditional general text classification, recruitment texts are often structured into distinct sections such as Title, Responsibilities, Requirements, and Skills, and each section conveys experience clues from different dimensions, transforming this task into a complex text comprehension problem involving multi-source information fusion.

Existing research indicates that job titles exhibit high ambiguity and inconsistency in practical use: the same title may correspond to different scopes of responsibility and qualification levels across industries and companies (Laosaengpha et al., 2024; Senger et al., 2024). Therefore, job titles alone cannot accurately indicate experience levels; a comprehensive assessment must be made by considering semantic information such as detailed responsibilities and qualification requirements within the job description. (Li et al., 2020). Driven by this challenge, relevant research has evolved from early rule-based systems to classical text feature methods like TF-IDF, then to large-scale pre-trained language models such as BERT, and most recently to large language models (LLMs) to enhance the capture of experiential clues.

Nevertheless, existing methods primarily treat job descriptions as a single continuous text, falling short of fully leveraging the differentiated information embedded within their paragraph structures: For instance, the "Responsibilities" section often reflects the complexity of tasks and scope of the role, while the "Requirements" section typically contains the most direct indicators of qualifications, such as years of experience and skill levels. Thus, different paragraphs play unequal roles in determining experience levels. Therefore, the traditional practice of encoding entire job descriptions may weaken key clues and limit the model's ability to distinguish between candidates.

Based on this observation, this study proposes the Section-Aware BERT as the main model, which encodes recruitment texts by structured paragraphs and performs high-level feature fusion. This method enables the model to automatically focus on text segments with stronger experience indicators during the decision-making process, thereby enhancing sensitivity and discrimination capabilities for qualification clues. Furthermore, in order to comprehensively evaluate different modeling paradigms on this task, we systematically test large language models (LLMs) under both few-shot and fine-tuning supervision settings on the same dataset. We compare these against traditional models to explore the performance ceilings and applicable scenarios of different approaches.

To further understand model differences and limitations, we conducted both qualitative and quantitative error analysis, including confusion matrix comparisons, keyword extraction from misclassified samples as well as behavioral differences in all models. Results indicate that errors in experience level prediction exhibit high systematicity rather than randomness: significant confusion exists between entry-level and senior positions, while mid-level roles demonstrate persistent classification instability due to ambiguous semantic boundaries.

Overall, this study explores the job experience level classification task under relatively limited data conditions. This is achieved by constructing multiple baseline models, proposing a structured section-aware modeling approach, and combining few-shot and fine-tuning experiments with LLMs. We hope the analytical framework and model design proposed in this research will provide reference and inspiration for future studies in recruitment text processing.

The remainder of the paper is organized as follows: Section 2 reviews relevant literature; Section 3 describes the dataset and the preprocessing workflow; Section 4 illustrates baseline methods, including a rule-based regular expression system as well as a TF--IDF classifier; Section 5 details the advanced models including Section-Aware BERT and LLM models for both few shot and fine-tune; Section 6 presents the experimental results for evaluation framework; Section 7 presents quantitative and qualitative error analysis; Section 8 discusses the overall findings and limitations, and Section 9 concludes the paper and proposes future research directions.

\section{Related Work}\label{related-works}

Automatic analysis of recruitment texts has long been a core challenge in e-recruitment systems, with research primarily focused on job classification, position standardization, and the automated identification of competency as well as qualification information. Early approaches typically treated job advertisements as text classification problems, relying on manually designed features and traditional machine learning models to achieve efficient organization and management of large-scale recruitment data.

\citet{Malherbe2015} proposed an efficient job advertisement classification system to map online recruitment information to standardized job classification systems. This research employed lightweight text features based on word n-grams combined with traditional classification models, emphasizing computational efficiency and scalability while maintaining high classification accuracy. Such approaches represented the mainstream paradigm in recruitment text analysis prior to the advent of Transformer-based models, relying primarily on lexical-level cues and domain-knowledge-driven feature engineering. This research demonstrates that even in complex recruitment text scenarios, classic TF-IDF retains a strong baseline value for job classification.

Early efforts primarily relied on TF-IDF features and rule-based systems for classifying skills and job positions; Recent studies, however, increasingly adopt Transformer to capture richer semantic relationships within recruitment texts. With the advancement of deep learning methods, research focus has shifted from shallow textual features to semantic representation learning. \citet{Senger2024} conducted a systematic review of recent research in recruitment text analysis, noting the field's transition from keyword matching and feature engineering approaches to methods based on pre-trained language models. 

Within this broader shift from rule-based system toward Transformer-based modeling, several studies have specifically targeted experience-related inference tasks. For the specific task of experience level prediction, \citet{Li2020} proposed a context-aware Transformer-based competency level prediction model. By integrating document segmentation with multi-head attention mechanisms, they performed joint modeling of resumes and job descriptions. Using high-quality manually annotated data, this study validated the effectiveness of Transformer models in experience level prediction, achieving approximately 73\% classification accuracy. Although their research primarily focused on candidate resumes rather than job advertisements, the segmented modeling approach they proposed is highly relevant to the experience level prediction task.

Concurrently, the role of job titles in recruitment text analysis has also attracted significant attention from researchers. \citet{Laosaengpha2024} proposed a job title representation learning method based on job description aggregation to mitigate inconsistencies in job title usage across companies and industries. Their research indicates that job titles alone are insufficient to accurately capture job semantics; modeling must rely on complete job description texts. This finding further confirms that using only job titles as input is inadequate for experience level prediction tasks, requiring thorough extraction of information from paragraphs detailing responsibilities, requirements, and skills.

More broadly, prior research has long recognized the importance of exploiting the internal structure of recruitment texts. Such texts typically feature a clear and diverse internal organizational structure, with different paragraphs serving distinct semantic functions. \citet{GnehmClematide2020} proposed a two-stage recruitment text processing framework encompassing text zoning and hierarchical classification, achieving high-precision paragraph identification using a BiLSTM-CRF model. Subsequently, they employed BERT for multidimensional job classification on zoned text, validating the effectiveness of structured processing in recruitment text understanding. This study demonstrates that explicitly leveraging document structure when modeling recruitment texts helps models more accurately locate task-relevant information.

\section{Dataset and Preprocessing}\label{dataset-and-preprocessing}

This section describes the dataset used in our experiments and the preprocessing steps applied prior to model training, including experience level mapping (section 3.2), data cleaning (section 3.3), section segmentation (section 3.4), and dataset splitting (section 3.5).

\subsection{Dataset Description}\label{dataset-description}

The data used in this study is sourced from the publicly available
LinkedIn Job Postings Dataset on Kaggle \citep{Arshkon2023}. This dataset comprises approximately 124,000 job postings annotated with experience level information. Experience level information is voluntarily provided by recruiters when posting jobs on LinkedIn, reflecting the qualification requirements for the position. Therefore, it can be considered a relatively reliable source of supervisory signals.

\subsection{Experience Level Mapping}

The original dataset contained six distinct experience level categories. However, due to variations in how different companies and industries define experience levels, directly using the original labels may introduce label noise and inconsistency. Thus, in order to enhance label consistency and maintain compatibility with prior research, we followed the approach of \citet{Li2020} and mapped the original six categories to four standardized categories. The mapping scheme is as follows: "Entry-level" and "Internship" are both mapped to Entry; "Associate" is mapped to Mid; "Mid-Senior level" is mapped to Senior; "Director'' and "Executive" remain Executive-level designation.

\subsection{Data Cleaning}\label{data-cleaning}

The original LinkedIn recruitment data exhibits significant variations
in text formatting and content quality due to diverse sources, including
missing fields, duplicate job postings, inconsistent text structures,
and extremely short or invalid descriptions. To ensure the model learns
semantic features for experience level prediction from high-quality
samples, this study implemented a systematic data cleaning process.

First, we retained only the four fields most relevant to experience
level prediction, labeled as \emph{title, description, skills\_desc,}
and \emph{formatted\_experience\_level}; After that, we removed samples
with missing descriptions and filtered out job postings with
descriptions shorter than 30 characters to eliminate noise from empty or
templated text. Completely duplicate description texts were removed to
prevent model bias toward repetitive samples during training as well.

When handling the \emph{skills\_desc} field, we chose to replace missing
values with empty strings, as skill information is optional across
different positions. Discarding these samples outright would cause data
distribution skew; However, since the experience level label
(\emph{formatted\_experience\_level}) is an essential annotation
required for supervised learning, missing samples in this field were
directly discarded.

Last, we standardized the text format by removing special characters and
merging redundant whitespace and line breaks in order to enhance the
stability of subsequent representation learning.

\subsection{Section Segmentation}\label{section-segmentation}

After completing data cleaning and merging experience level labels, we
ultimately obtained 80,717 job postings data with valid experience level
annotations. The distribution of experience levels is as follows:
Senior-level accounts for 36,110 entries, Entry-level for 31,453
entries, Mid-level for 8,719 entries, and Executive-level for 4,435
entries.

We then performed a paragraph-level structural analysis on each job
posting, automatically extracting three text segments: Responsibilities,
Requirements, and Other. The segmentation results are summarized as
follows:

\begin{itemize}
\item
 Responsibilities paragraph extraction success rate: 97.9\% (79,023 /
 80,717)
\item
 Requirements paragraph extraction success rate: 91.1\% (73,495 /
 80,717)
\item
 Other paragraphs were obtainable in all samples (100\%)
\end{itemize}

For texts where paragraph headings cannot be clearly identified, the system supplements information through alternative strategies such text segmentation, thereby minimizing information loss.

\subsection{Dataset Splits}\label{dataset-splits}

To ensure consistent class distributions during model training and
evaluation, we employ stratified sampling based on experience-level
labels to partition the dataset into training, development, and test
sets. Specifically, the dataset is divided as follows:

\begin{itemize}
\item
 Training set: 56,501 instances (70\%)
\item
 Development set: 12,108 instances (15\%)
\item
 Test set: 12,108 instances (15\%)
\end{itemize}

This splitting strategy ensures that minority categories, such as
Executive-level, are adequately represented across all subsets, thereby
mitigating assessment biases that may arise from underrepresentation or
significant imbalances within categories.

\begin{table}[t]
\centering
\small
\renewcommand{\arraystretch}{1.2}
\begin{tabular}{
  p{0.12\linewidth}
  p{0.38\linewidth}
  p{0.38\linewidth}
}
\toprule
\textbf{Level} &
\textbf{Title Patterns} &
\textbf{Description Patterns} \\
\midrule
Executive &
director, vice president, vp, chief, ceo, cto, cfo, head of, president &
reports to ceo/board, leads team of 10+, p\&l responsibility \\
\addlinespace
Entry &
entry-level, junior, intern, trainee, apprentice &
no experience required, 0--2 years, recent graduate, will train \\
\addlinespace
Senior &
senior, sr., lead, principal, staff, architect &
5+ years experience, 10+ years, extensive experience, proven track record \\
\addlinespace
Mid &
associate (excl.\ senior associate) &
2--5 years experience, some experience \\
\bottomrule
\end{tabular}
\caption{Regular expression patterns for rule-based classification.}
\label{tab:regex-patterns}
\end{table}

\section{Baseline Methods}\label{sec:baseline-methods}
This section presents the baseline methods used for comparison in this study, including a rule-based regular expression system (Section 4.1) and a classical machine learning model based on TF-IDF features and logistic regression (Section 4.2).

\subsection{Rule-Based BaselineRegex System}\label{sec:rule-based-regex}

As a simple baseline, we implemented a rule-based classification system
using regular expressions. This approach relies on handcrafted patterns
that match explicit experience-level indicators in job titles and descriptions.

The system applies patterns in a hierarchical order, checking for
Executive-level indicators first, followed by Entry, Mid, and Senior
levels. Table~\ref{tab:regex-patterns} summarizes the key patterns used for each category.

The classification logic proceeds as follows: First, the system checks
the job title against Executive-level patterns (e.g., "director," "vice
president," "chief"). If no match is found, it checks for Entry-level
patterns (e.g., "junior," "intern," "entry-level"). The system then
checks for the "associate" keyword to identify Mid-level positions,
excluding cases where "senior associate" appears. Finally, it checks for
Senior-level patterns (e.g., "senior," "lead," "principal").

If no title patterns match, the system examines the job description for
experience-related phrases. Patterns such as "no experience required" or
"0-2 years" indicate Entry level, while "5+ years experience" or "proven
track record" indicate Senior level. If no patterns match at any stage,
the system defaults to Senior, as this is the most frequent category in
the dataset.

\subsection{Classical ML Baseline: TF-IDF + Logistic Regression}\label{subsec:tfidf-baseline}

As a classical machine learning baseline, we implemented a TF-IDF vectorizer combined with logistic regression. This approach learns statistical patterns from the training data rather than relying on handcrafted rules.

For each job posting, we concatenated the job title and description into a single text field. We then applied TF-IDF vectorization with a maximum of 5,000 features, using both unigrams and bigrams. We set the minimum document frequency to 5 and maximum document frequency to 80\% to filter out overly rare or common terms, and removed English stop words.

We used logistic regression with L2 regularization as the classifier. To select the optimal regularization strength, we performed hyperparameter tuning on the development set, testing C values of 0.01, 0.1, 1, 10, and 100. Based on development set performance, we selected C = 10 for the final model.

\section{Advanced Models}\label{sec:advanced-models}
This section introduces the advanced modeling approaches evaluated in our experiments, including Baseline BERT (Section~\ref{subsec:baseline-bert-model}), our proposed Section-Aware BERT model that explicitly incorporates document structure (Section~\ref{subsec:section-aware-bert-model}), and large language models (LLMs) used as ceiling baselines under different supervision settings (Section~\ref{subsec:llm-baseline}).

\subsection{Baseline BERT}\label{subsec:baseline-bert-model}

\begin{figure}[h]
  \centering
  \includegraphics[width=0.5\columnwidth]{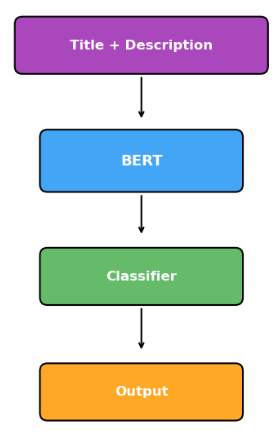}
  \caption{Baseline BERT architecture. The concatenated title and description are encoded as a single sequence.}
  \label{fig:baseline-bert-arch}
\end{figure}

As a strong neural baseline, we fine-tune BERT-base-uncased on the job classification task. For each job posting, we concatenate the job title and description into a single text sequence, which is tokenized and truncated to a maximum length of 512 tokens. The [CLS] token representation is passed through a dropout layer (p = 0.1) and a linear classifier to produce the four-class prediction (Figure~\ref{fig:baseline-bert-arch}).

We use the AdamW optimizer with a learning rate of 2e-5 and a linear warmup schedule covering 10\% of total training steps. The model is trained for 3 epochs with a batch size of 16, and we select the best checkpoint based on development set performance.

While Baseline BERT captures contextual semantics that bag-of-words models miss, it treats the entire job posting as an undifferentiated sequence. The job title---despite being a strong experience signal---occupies only a small fraction of the 512-token input and may be diluted by lengthy descriptions.

\subsection{Section-Aware BERT}\label{subsec:section-aware-bert-model}

\begin{figure}[h]
    \centering
    \includegraphics[width=0.75\columnwidth]{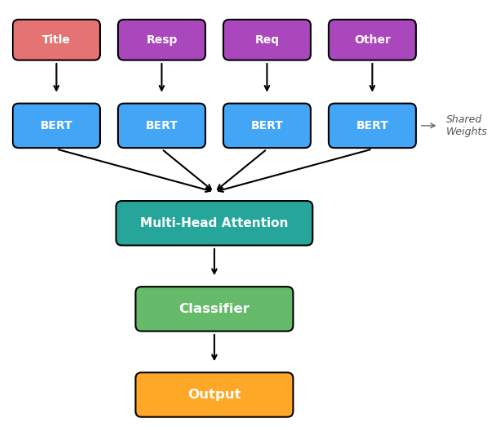}
    \caption{Section-Aware BERT architecture. Four sections (Title, Responsibilities, Requirements, Other) are encoded separately by a shared BERT encoder, then fused via multi-head attention.}
    \label{fig:section-aware-arch}
\end{figure}

As discussed in Introduction, job descriptions typically contain distinct sections that carry different signals for experience level prediction. The job title often contains explicit seniority markers; the Responsibilities section reflects task complexity and scope; while the Requirements section contains explicit qualification indicators such as years of experience. To leverage this structure, we propose Section-Aware BERT, which encodes each section separately before fusing them with multi-head attention (Figure~\ref{fig:section-aware-arch}).

\subsubsection{Model Architecture}

The model architecture consists of three components: a shared BERT encoder, a multi-head attention fusion layer, and a classification head.

Given a job posting, each of the four sections is independently encoded using a shared BERT-base-uncased encoder. We extract the [CLS] token representation from each section, yielding four 768-dimensional vectors. Formally, let $h_{\text{title}}, h_{\text{resp}}, h_{\text{req}}, h_{\text{other}} \in \mathbb{R}^{768}$ denote the representations for Title, Responsibilities, Requirements, and Other sections respectively. We stack these vectors to form $H \in \mathbb{R}^{4 \times 768}$.

To learn the relative importance of each section, we apply an 8-head self-attention mechanism:
\begin{equation}
A = \text{MultiHeadAttention}(H, H, H)
\end{equation}

We then apply a residual connection followed by layer normalization:
\begin{equation}
H' = \text{LayerNorm}(H + A)
\end{equation}

The fused representation is obtained by mean pooling over the four attended section vectors, which is then passed through a dropout layer (p = 0.1) and a linear classifier to produce the final four-class prediction.

\subsubsection{Training}

We use the same training configuration as Baseline BERT: AdamW optimizer with learning rate 2e-5, linear warmup covering 10\% of training steps, 3 epochs, and batch size 16. All four sections share the same BERT encoder, which reduces parameters and encourages generalizable representations across sections.

Note that Section-Aware BERT requires four forward passes through BERT per sample (one per section), compared to one pass for Baseline BERT. However, the shorter sequence lengths (64--256 tokens per section vs. 512 tokens) partially offset this computational cost.

\subsection{LLM Baseline}\label{subsec:llm-baseline}

To compare different modeling paradigms on this task, we evaluated GPT-4o-mini under two settings: few-shot prompting and fine-tuning. This comparison allows us to examine whether general-purpose LLMs can match task-specific architectures that leverage document structure.

\subsubsection{Few-Shot LLM}
For the few-shot setting, we designed a prompt that includes task instructions, label definitions, and demonstration examples. The system prompt explicitly defines the four experience levels according to LinkedIn\textquotesingle{}s labeling conventions, noting that LinkedIn\textquotesingle{}s "Mid-Senior level" maps to our Senior category, and that "Entry level" includes internships.

We constructed 12 few-shot examples, with three examples per class, carefully selected to cover diverse job types and industries. Each example includes a job title, a truncated description, and the correct classification. The examples were chosen to illustrate typical patterns for each level: Entry-level examples include assistants, analysts, and technicians; Mid-level examples include engineers and accountants with moderate experience; Senior-level examples include managers and leads; Executive-level examples include directors, controllers, and C-suite positions.

For inference, we concatenate the system prompt, few-shot examples, and the test job posting into a single prompt. The job description is truncated to 1,500 characters to fit within context limits. The model is instructed to respond with only the category name. We parse the response by checking for the presence of each label and default to Senior (the most frequent class) if no valid label is detected.

Due to API cost constraints, we evaluated on a stratified sample of 400 test examples, maintaining the original class distribution.

\begin{table*}[!t]
\centering
\small
\renewcommand{\arraystretch}{1.15}
\begin{tabular*}{\textwidth}{@{\extracolsep{\fill}}lcccc@{}}
\toprule
\textbf{Model} & \textbf{Accuracy} & \textbf{Macro F1} & \textbf{Weighted F1} & \textbf{Test Samples} \\
\midrule
\multicolumn{5}{l}{\textit{Classical \& Rule-Based Models}} \\
Rule-Based (Regex) & 0.5035 & 0.4284 & 0.4160 & 12,108 \\
TF-IDF + Logistic Regression & 0.7341 & 0.6351 & 0.7202 & 12,108 \\
\midrule
\multicolumn{5}{l}{\textit{Transformer-Based Models}} \\
Baseline BERT & 0.7835 & 0.7200 & 0.7807 & 12,108 \\
\textbf{Section-Aware BERT} & \textbf{0.8256} & \textbf{0.7600} & \textbf{0.8200} & 12,108 \\
\midrule
\multicolumn{5}{l}{\textit{Large Language Models (LLMs)}} \\
GPT-4o-mini (few-shot) & 0.6432 & 0.6025 & 0.6440 & 398 \\
GPT-4o-mini (fine-tuned) & 0.7186 & 0.6108 & 0.6935 & 398 \\
\bottomrule
\end{tabular*}
\caption{Overall performance comparison across classical, transformer, and LLM-based models. For GPT-4o-mini models, evaluation is conducted on a stratified subset of 398 test samples due to API constraints.}
\label{tab:overall}
\end{table*}

\subsubsection{Fine-Tuned LLM}
For the fine-tuning setting, we used OpenAI\textquotesingle{}s fine-tuning API to adapt GPT-4o-mini to our specific task. We prepared training data in the chat completion format, with each example consisting of a system message defining the task, a user message containing the job title and description, and an assistant message containing the correct label.

We created a training set of 1,000 examples and a validation set of 200 examples, both using stratified sampling to preserve class distribution. Note that this represents only 1.8\% of the training data used for BERT models (1,000 vs.\ 56,501 examples), which limits the direct comparability of results. Due to API cost constraints, scaling to the full training set was not feasible for this study. The job descriptions were truncated to 1,500 characters.
The system prompt for fine-tuning was simplified compared to the few-shot setting: "Classify the job posting into exactly one category: Entry, Mid, Senior, or Executive. Respond with only the category name."

The fine-tuning process was conducted using the GPT-4o-mini-2024-07-18 model via OpenAI's API. After training completed, we evaluated the fine-tuned model on the same stratified sample of 400 test examples used for the few-shot evaluation, enabling direct comparison between the two settings.

\section{Results}\label{results}

This section presents the experimental results comparing all proposed
models on the job experience level prediction task. We evaluate
performance using accuracy, macro-averaged F1 score, and weighted F1 score to account for class imbalance in the dataset. Note that for GPT-4o-mini models, due to API cost and computational constraints, we evaluated on a stratified random subset of 398 test samples rather than the entire 12,108-sample test set.

\subsection{Overall Performance
Comparison}\label{overall-performance-comparison}

Table \ref{tab:overall} presents the overall performance of all models. The results
demonstrate clear performance hierarchies across different modeling
paradigms, with notable improvements from rule-based methods to
transformer-based approaches.

The rule-based regex system achieves 50.35\% accuracy, establishing a
baseline that relies on handcrafted patterns. While it captures explicit
signals like "director" or "intern" in job titles, it struggles with
implicit experience indicators.

The TF-IDF + Logistic Regression model shows substantial improvement,
achieving 73.41\% accuracy (23 percentage points over regex),
demonstrating that statistical learning outperforms handcrafted rules.

Section-Aware BERT achieves the strongest performance with 82.56\%
accuracy and 76.00\% macro F1, representing a 9.15 percentage point
improvement over TF-IDF. The model encodes job postings by structured
sections (Title, Responsibilities, Requirements, and Other) and uses
multi-head attention to learn their relative importance. This
architecture effectively captures both explicit qualification indicators
and implicit signals about task complexity across multiple document
sections.

GPT-4o-mini in few-shot mode achieves 64.32\% accuracy, outperforming the regex baseline but underperforming classical TF-IDF. The performance gap is attributed to the disparity in size of training data. However, it also reveals a broader finding: task-specific structural inductive biases can be more effective than increased model capacity alone. Despite having notably fewer parameters (110M vs. GPT-4o-mini's larger architecture), Section-Aware BERT's explicit section-level modeling outperforms the general-purpose LLM, suggesting that for structured text classification tasks, domain-appropriate architectural design may yield greater returns than scale.

\begin{table*}[!t]
\centering
\small
\renewcommand{\arraystretch}{1.15}
\begin{tabular*}{\textwidth}{@{\extracolsep{\fill}}lccccc@{}}
\toprule
\textbf{Model} & \textbf{Entry} & \textbf{Mid} & \textbf{Senior} & \textbf{Executive} & \textbf{Macro Avg} \\
\midrule
\multicolumn{6}{l}{\textit{Classical \& Rule-Based Models}} \\
Rule-Based (Regex) & 0.18 & 0.13 & 0.65 & 0.75 & 0.43 \\
TF-IDF + Logistic Regression & 0.78 & 0.36 & 0.77 & 0.63 & 0.64 \\
\midrule
\multicolumn{6}{l}{\textit{Transformer-Based Models}} \\
Baseline BERT & 0.83 & 0.53 & 0.80 & 0.73 & 0.72 \\
\textbf{Section-Aware BERT} & \textbf{0.88} & \textbf{0.54} & \textbf{0.85} & 0.79 & \textbf{0.76} \\
\midrule
\multicolumn{6}{l}{\textit{Large Language Models (LLMs)}} \\
GPT-4o-mini (few-shot) & 0.71 & 0.18 & 0.68 & \textbf{0.85} & 0.60 \\
GPT-4o-mini (fine-tuned) & 0.77 & 0.11 & 0.76 & 0.81 & 0.61 \\
\bottomrule
\end{tabular*}
\caption{Per-class F1 scores for each model on the test set.}
\label{tab:perclass}
\end{table*}

\subsection{Per-Class Performance Analysis}

Table~\ref{tab:perclass} presents the per-class F1-scores for all models, providing a comprehensive comparison across experience levels.

The regex system shows highly imbalanced performance, with extremely
poor results for Entry (0.18) and Mid (0.13) levels due to its
default-to-Senior fallback behavior. It performs better on Executive
(0.75) due to explicit title patterns.

TF-IDF + Logistic Regression demonstrates notable improvements
across Entry (0.78) and Senior (0.77) levels, with a 48\% relative
improvement in macro F1 over regex. However, Mid-level remains weak
(0.36), and Executive recall is limited (0.63).

Section-Aware BERT achieves the strongest performance across all
categories, with particularly impressive results for Entry (0.88) and
Senior (0.85). The model shows consistent improvements over TF-IDF:
+0.10 for Entry, +0.18 for Mid, +0.08 for Senior, and +0.16 for
Executive. Mid-level (0.54) remains the most challenging category but
shows substantial
gains over all baselines.

GPT-4o-mini in few-shot mode performs well on Executive positions (0.85) but struggles significantly with Mid-level (0.18). Fine-tuning improves Entry and Senior classification but causes Mid-level recall to collapse (0.11), as the model becomes overly conservative for this ambiguous category with limited training samples.

\section{Error Analysis}\label{sec:error-analysis}

This section provides a detailed error analysis across all evaluated models, examining how each modeling paradigm addresses the limitations of its predecessors. We analyze the progression from rule-based systems to transformer-based approaches, highlighting the specific error patterns and linguistic challenges at each stage.

\subsection{Rule-Based System Limitations}\label{subsec:regex-errors}

The rule-based regex system achieves only 50.4\% accuracy due to three fundamental limitations.

\textbf{Low Keyword Coverage.} 
Only 24\% of test samples contain explicit experience-level keywords. Entry-level positions show merely 4.7\% coverage, as most lack explicit markers like "junior" or "intern," while Executive positions achieve 70.9\% coverage due to distinctive titles.

\textbf{Default-to-Senior Bias.} 
The 76\% of samples without keywords default to Senior (the majority class), causing 4,047 Entry-level positions to be misclassified---the largest source of errors.

\textbf{Lexical Ambiguity.} 
Polysemous terms like "coordinator" (38.6\% error rate) and "analyst" (37.6\%) appear across all experience levels. The system cannot distinguish "Sales Coordinator" (Entry) from ``Project Coordinator'' (Mid) based on surface patterns alone.

These limitations motivate \textit{statistical}\textit{ learning approaches}: TF-IDF can learn implicit lexical associations from training data rather than relying on handcrafted keyword lists, potentially capturing experience signals that lack explicit markers.

\subsection{TF-IDF Vocabulary Limitations}\label{subsec:tfidf-errors}

TF-IDF with Logistic Regression achieves 73.5\% accuracy, a 23-percentage-point improvement over the regex baseline. However, its bag-of-words representation introduces fundamental limitations.

\textbf{Loss of Word Order.}
TF-IDF represents documents as unordered frequency vectors, discarding sequential information entirely. As illustrated in Table~\ref{tab:tfidf-wordorder}, sentences with opposite meanings produce identical representations, causing systematic confusion.

\begin{table}[!ht]
\centering
\small
\renewcommand{\arraystretch}{1.2}
\begin{tabular*}{\columnwidth}{@{\extracolsep{\fill}}p{0.45\columnwidth}p{0.45\columnwidth}@{}}
\toprule
\textbf{Sentence A} & \textbf{Sentence B} \\
\midrule
``you will lead the team'' & ``the team will lead you'' \\
\addlinespace
``entry-level management training'' & ``management of entry-level staff'' \\
\addlinespace
``assist senior engineers'' & ``senior engineer assists'' \\
\bottomrule
\end{tabular*}
\caption{Word order examples that TF-IDF cannot distinguish. Each pair implies different seniority levels but produces identical bag-of-words vectors.}
\label{tab:tfidf-wordorder}
\end{table}

\textbf{Polysemy and Contextual Blindness.}
TF-IDF assigns each word a single global weight, ignoring context-dependent meanings. As shown in Table~\ref{tab:tfidf-ambiguity}, polysemous terms exhibit high error rates: "coordinator" (38.6\%), "analyst" (37.6\%), and "specialist" (35.2\%). The word "lead" alone illustrates this problem---it may indicate leadership responsibility (Senior) or appear in "lead generation" (Entry-level marketing). Without contextual understanding, TF-IDF cannot disambiguate these usages.

\begin{table}[h]
\centering
\small
\renewcommand{\arraystretch}{1.15}
\begin{tabular*}{\columnwidth}{@{\extracolsep{\fill}}lrrr@{}}
\toprule
\textbf{Term} & \textbf{Total} & \textbf{Errors} & \textbf{Error Rate} \\
\midrule
coordinator & 324   & 125 & 38.6\% \\
analyst     & 529   & 199 & 37.6\% \\
specialist  & 642   & 226 & 35.2\% \\
engineer    & 1,080 & 352 & 32.6\% \\
associate   & 582   & 166 & 28.5\% \\
\bottomrule
\end{tabular*}
\caption{TF-IDF error rates for polysemous job title terms.}
\label{tab:tfidf-ambiguity}
\end{table}

\begin{figure}[t]
    \centering
    \includegraphics[width=0.95\columnwidth]{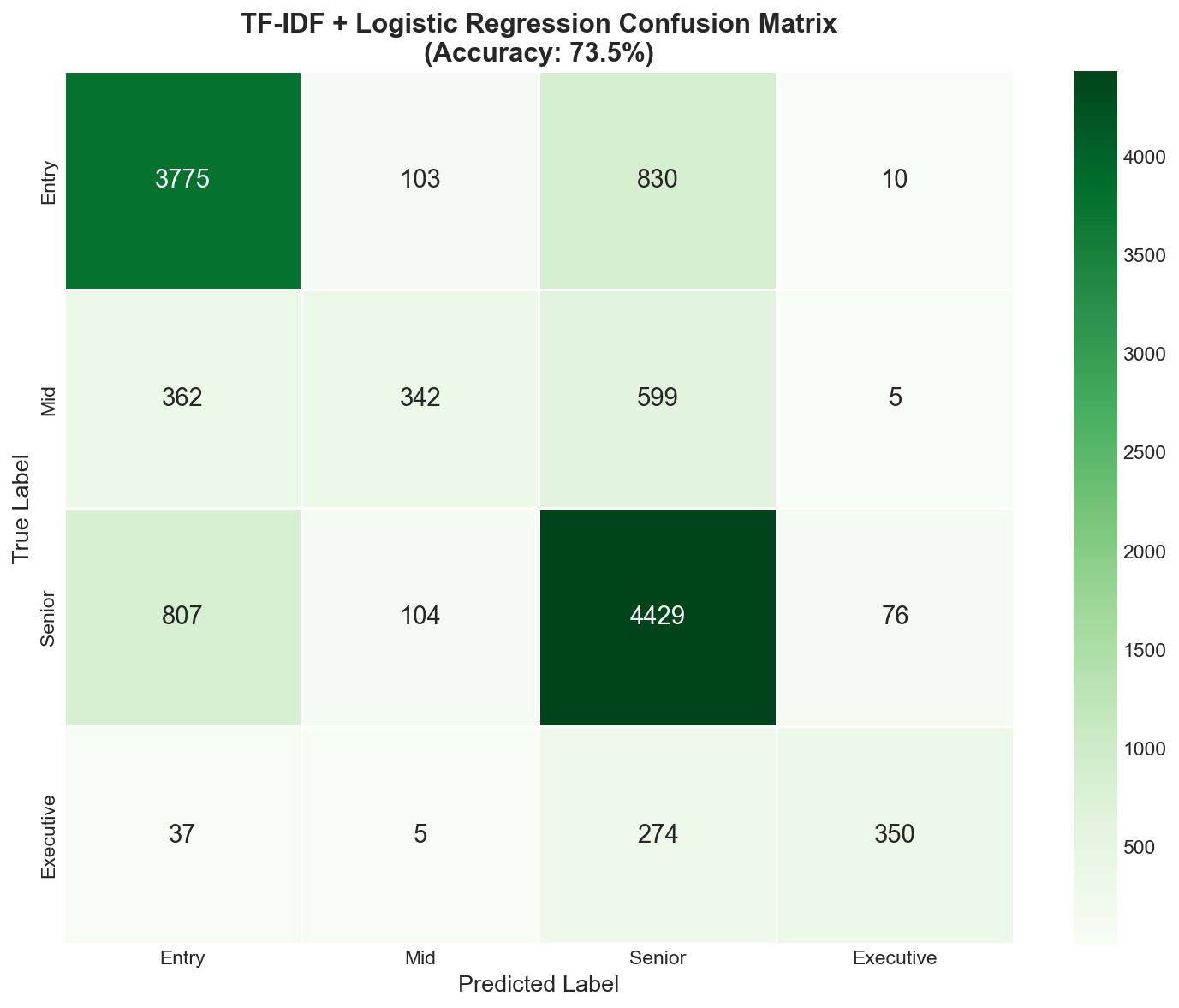}
    \caption{TF-IDF confusion matrix. Bidirectional Entry$\leftrightarrow$Senior confusion dominates due to shared vocabulary across domains.}
    \label{fig:tfidf-cm}
\end{figure}

\textbf{Confusion Patterns.}
Figure~\ref{fig:tfidf-cm} shows that Entry$\rightarrow$Senior (830 errors) and Senior$\rightarrow$Entry (807 errors) dominate, accounting for over 50\% of all errors. Mid-level remains the hardest category with only 36.7\% F1, as its vocabulary overlaps with both Entry ("support," "assist") and Senior ("manage," "lead").

These limitations motivate \textit{transformer-based models}: BERT learns contextualized embeddings where the same word receives different representations based on surrounding context, enabling disambiguation of polysemous terms and sensitivity to word order.

\subsection{BERT Model Comparison}\label{subsec:bert-errors}

We evaluate transformer-based approaches that address TF-IDF's contextual limitations.

\subsubsection{Baseline BERT}\label{subsubsec:baseline-bert}

Baseline BERT encodes the concatenated job title and description as a single sequence (maximum 512 tokens), achieving 78.5\% accuracy---a 5-percentage-point improvement over TF-IDF. The contextualized embeddings successfully disambiguate polysemous terms by attending to surrounding context. However, the job title occupies only a small fraction of the input and may be diluted by lengthy descriptions.

\subsubsection{Section-Aware BERT}\label{subsubsec:section-aware}

Section-Aware BERT encodes job postings as four separate sections: Title (64 tokens), Responsibilities (256 tokens), Requirements (256 tokens), and Other (256 tokens). A multi-head attention mechanism learns the relative importance of each section before classification.

This architecture yields 82.6\% accuracy, a 4.1-percentage-point improvement over Baseline BERT. As shown in Table~\ref{tab:bert-comparison}, all classes benefit from explicit section modeling, with Executive showing the largest gain (+8.1\%).

\begin{table}[!ht]
\centering
\small
\renewcommand{\arraystretch}{1.15}
\begin{tabular*}{\columnwidth}{@{\extracolsep{\fill}}lrrr@{}}
\toprule
\textbf{Class} & \textbf{Baseline} & \textbf{Section-Aware} & \textbf{$\Delta$} \\
\midrule
Entry     & 83.3\% & 87.1\% & +3.8\% \\
Mid       & 47.2\% & 50.6\% & +3.4\% \\
Senior    & 83.0\% & 86.9\% & +3.9\% \\
Executive & 69.5\% & 77.6\% & +8.1\% \\
\midrule
\textbf{Overall} & \textbf{78.5\%} & \textbf{82.6\%} & \textbf{+4.1\%} \\
\bottomrule
\end{tabular*}
\caption{Recall comparison between Baseline BERT and Section-Aware BERT.}
\label{tab:bert-comparison}
\end{table}

\begin{figure}[h]
    \centering
    \includegraphics[width=0.90\columnwidth]{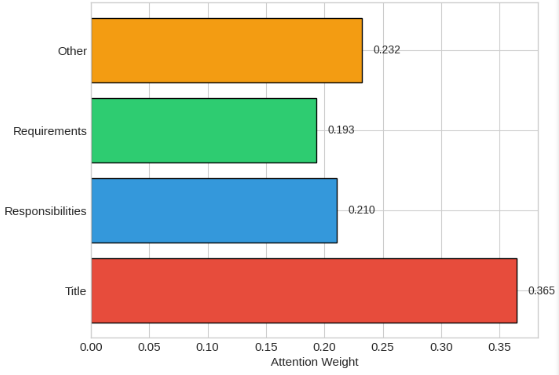}
    \caption{Section attention weights in Section-Aware BERT. Title receives the highest attention (36.5\%), while Responsibilities and Requirements together contribute 40.3\%.}
    \label{fig:attention-weights}
\end{figure}

\textbf{Attention Weight Analysis.}
The learned attention weights reveal which sections the model prioritizes. As shown in Figure~\ref{fig:attention-weights}, Title receives the highest average attention (36.5\%), followed by Other (23.2\%), Responsibilities (21.0\%), and Requirements (19.3\%). This confirms that while job titles carry the strongest signal, all sections contribute to the final prediction.

\textbf{Section-Specific Signal Analysis.}
To understand the contribution of each section beyond Title, we analyzed the experience signals present in Responsibilities and Requirements. Table~\ref{tab:section-signals} shows that these sections contain distinct indicators: Requirements captures explicit qualifications (years of experience, degree requirements), while Responsibilities captures implicit seniority through task complexity and scope.

\begin{table}[h]
\centering
\small
\begin{tabular}{lr|lr}
\toprule
\multicolumn{2}{c|}{\textbf{Requirements}} & \multicolumn{2}{c}{\textbf{Responsibilities}} \\
\midrule
Years mentioned & 35.5\% & Leadership verbs & 65.6\% \\
Degree required & 36.7\% & Strategic scope & 20.5\% \\
Certification & 28.6\% & Entry-level verbs & 69.1\% \\
\bottomrule
\end{tabular}
\caption{Experience signals by section (Section-Aware Wins samples).}
\label{tab:section-signals}
\end{table}

Notably, signal strength varies systematically by experience level. Table~\ref{tab:signal-by-level} shows that among samples where Section-Aware wins, Executive positions exhibit 81.2\% prevalence of leadership verbs and 40.0\% strategic scope indicators, compared to only 60.0\% and 15.0\% for Entry positions. This explains why Executive shows the largest improvement (+8.1\%): the Responsibilities section captures strategic and leadership signals that distinguish executives from senior individual contributors.

\begin{table}[!ht]
\centering
\small
\setlength{\tabcolsep}{3pt}
\renewcommand{\arraystretch}{1.15}
\begin{tabular*}{\columnwidth}{@{\extracolsep{\fill}}lrrrr@{}}
\toprule
\textbf{True Label} & \textbf{Leader.} & \textbf{Strat.} & \textbf{Years} & \textbf{EntryV.} \\
\midrule
Entry     & 60.0\% & 15.0\% & 32.4\% & 68.7\% \\
Mid       & 59.0\% & 13.5\% & 33.3\% & 65.4\% \\
Senior    & 60.6\% & 17.4\% & 35.4\% & 62.7\% \\
Executive & 81.2\% & 40.0\% & 55.0\% & 55.0\% \\
\bottomrule
\end{tabular*}
\caption{Signal prevalence (\%) by true experience level.}
\label{tab:signal-by-level}
\end{table}

\begin{table}[t]
\centering
\small
\renewcommand{\arraystretch}{1.15}
\begin{tabular*}{\columnwidth}{@{\extracolsep{\fill}}lrl@{}}
\toprule
\textbf{Confusion Fixed} & \textbf{Count} & \textbf{Example Title} \\
\midrule
Senior$\rightarrow$Entry & 323 & Customer Advocacy Mgr \\
Entry$\rightarrow$Senior & 296 & Process Engineer \\
Mid$\rightarrow$Senior & 80 & Orthopedics Distributor \\
Exec.$\rightarrow$Senior & 70 & Division Controller \\
\bottomrule
\end{tabular*}
\caption{Top confusion pairs fixed by Section-Aware BERT.}
\label{tab:confusion-fixed}
\end{table}

\textbf{Truncation Risk in Baseline BERT.}
A key architectural advantage of Section-Aware is avoiding truncation of important content. Analysis reveals that the Requirements section appears on average at 44.5\% into the description, with 29.9\% of samples having Requirements in the second half of the text. In Baseline BERT's 512-token limit (approximately 2,000 characters), descriptions averaging 3,900 characters face significant truncation risk. Section-Aware's dedicated 256 tokens per section ensures that experience signals in Requirements and Responsibilities are preserved regardless of their position in the original text.

\textbf{Confusion Patterns Fixed.}
We analyzed the 1,048 samples where Section-Aware succeeds but Baseline fails. As shown in Table~\ref{tab:confusion-fixed}, the model primarily resolves bidirectional Entry$\leftrightarrow$Senior confusion. These samples typically have ambiguous titles where experience level can only be determined from section content---such as years of experience in Requirements or task scope in Responsibilities.

\section{Discussion and Limitations}\label{sec:discussion}

\subsection{Strengths}\label{subsec:strengths}

Our study makes three key contributions. First, our novel approach Section-Aware BERT's explicit modeling of document structure achieves 82.6\% accuracy, outperforming TF-IDF (73.4\%) and fine-tuned LLMs (71.9\%), demonstrating the value of structured text processing. Second, our systematic evaluation across rule-based, TF-IDF, BERT, and LLM approaches establishes clear performance hierarchies and reveals limitations of each paradigm, providing insights for future recruitment text analysis. Third, our attention weight analysis provides interpretable insights, showing job titles contribute most to predictions (36\%), followed by responsibilities (21\%).

\subsection{Limitations}\label{subsec:limitations}

We note several limitations. First, experience-level labels on LinkedIn are employer-provided and can be noisy across industries and companies, which bounds achievable performance. Second, our rule-based section segmentation may introduce boundary noise when headings are missing or unconventional. Third, due to API cost constraints, LLM evaluations were conducted on a stratified subset rather than the full test set. Finally, the Mid-level category remains challenging across all models (best F1: 0.54), reflecting inherent ambiguity in this transitional tier.

\section{Conclusion and Future Work}\label{sec:conclusion}

This section concludes the paper by summarizing the main findings (Section~\ref{subsec:conclusion}) of the research and exploring potential directions for future studies (Section~\ref{subsec:future-work}).

\subsection{Conclusion}\label{subsec:conclusion}

This paper systematically explores the prediction of experience levels in recruitment texts and compares multiple modeling paradigms on the same dataset, including rule-based regular expression systems, the classic TF--IDF + logistic regression model, standard BERT, section-aware BERT with segmented modeling, and large language models (LLMs) using both few-shot and fine-tuned methods.

Experimental results indicate that traditional baseline methods retain some reference value for this task, yet their performance is significantly constrained by lexical ambiguity and contextual gaps. The TF--IDF model exhibits substantial confusion between Entry--Senior and Mid-level categories, reflecting the difficulty of bag-of-words representations in capturing fine-grained semantic distinctions related to experience levels.

In contrast, by explicitly modeling the internal structure of recruitment texts, segment-aware BERT outperforms traditional methods in both overall performance and category balance. Particularly in scenarios with ambiguous job titles, section-based modeling more effectively leverages experience clues distributed across key paragraphs of the job description. After optimizing model architecture and training strategies, the final Section-Aware BERT achieves an accuracy of 82.5\%, delivering the best performance among all non-LLM models.

Further error analysis revealed complementary patterns in information utilization: Standard BERT excelled when explicit experiential indicators were present; while section-aware BERT demonstrated superiority in extracting and integrating implicit experiential information dispersed across multiple paragraphs.

\subsection{Future Work}\label{subsec:future-work}

Although the method proposed in this paper achieves significant improvements in the task of predicting experience level, several directions require further exploration.

First, further strengthening the modeling approach for job titles.

Both experimental results and error analysis indicate that job titles remain a valuable source of information for predicting experience levels. Although this study explicitly incorporated job titles as input into the segment-aware BERT architecture and co-modeled them with paragraphs detailing job responsibilities, thereby mitigating to some extent the dilution of job information during the structured segmentation process, analysis indicates that significant differences persist in the experiential cues conveyed by job titles across different industries and positions. Thus, future refinement could build upon this foundation to explore more refined title modeling approaches. For instance, introducing explicit interaction mechanisms between titles and body paragraphs which can more effectively balance explicit title cues with implicit experience signals embedded within paragraphs.

Second, explore more refined or flexible text segmentation and structural modeling methods.

The rule-based segmentation method employed in this study demonstrates good scalability on large-scale datasets, yet it may still introduce boundary noise. Future work could incorporate learning-based segmentation models (such as sequence labeling) or adopt dynamic attention mechanisms, enabling the model to adaptively determine paragraph weights across different task phases and thereby reduce reliance on fixed segmentation rules.

Besides these two points, future research could also explore larger-scale or domain-adaptive pre-trained models, combined with multi-task learning or cross-dataset transfer learning, to mitigate noise stemming from inconsistent label definitions across recruitment platforms.

Overall, this study demonstrates that structure-aware text modeling and explicit modeling of experiential cues play a crucial role in enhancing recruitment text comprehension and predicting experience levels. We hope the analytical framework and model design proposed in this paper will provide valuable reference for future research on structured recruitment text understanding.

\bibliography{references}

\end{document}